# LLM Prompt Evaluation for Educational Applications*

Langdon Holmes[1,*], Adam Coscia[2], Scott Crossley[1], Joon Suh Choi[1], and Wesley Morris[1]

[1] *Vanderbilt University, Nashville, Tennessee*
[2] *Georgia Institute of Technology, Atlanta, Georgia*

**Abstract**

As large language models (LLMs) become increasingly common in educational applications, there is a growing need for evidence-based methods to design and evaluate LLM prompts that produce personalized and pedagogically aligned out-puts. This study presents a generalizable, systematic approach for evaluating prompts, demonstrated through an analysis of LLM-generated follow-up questions in a structured dialogue activity. Six prompt templates were designed and tested. The templates incorporated established prompt engineering patterns, with each prompt emphasizing distinct pedagogical strategies. The prompt templates were compared through a tournament-style evaluation framework that can be adapted for other educational applications. The tournament employed the Glicko2 rating system with eight judges evaluating question pairs across three dimensions: format, dialogue support, and appropriateness for learners. Data was sourced from 120 authentic user interactions across three distinct educational deployments. Results showed that a single prompt related to strategic reading out-performed other templates with win probabilities ranging from 81% to 100% in pairwise comparisons. This prompt combined persona and context manager pat-terns and was designed to support metacognitive learning strategies such as self-directed learning. The methodology showcases how educational technology re-searchers can systematically evaluate and improve prompt designs, moving be-yond ad-hoc prompt engineering toward evidence-based prompt development for educational applications.

## 1. Introduction

Large language models (LLMs) have become increasingly integral to educational platforms and applications. Even prior to the advent of large generative models such as the GPT (Generative Pre-trained Transformers) series [5], LLMs demonstrated success in various educational contexts with applications in automated writing evaluation [2, 25, 30] and textual analysis [35, 46]. These early applications primarily focused on specific, constrained tasks that relied on carefully trained models for particular educational use cases.

The advent of flexible and customizable LLMs, like GPTs and similar foundation models, has created an even greater demand for new and improved AI-powered applications in educational technology. Use cases now range from learner-facing chat bots that serve as study aids [3], sources of information on course content [41], and assistants for giving personalized feedback [22], to interactive systems for generating educational materials such as lesson plans and study guides [16]. For example, Taneja et al. developed Jill Watson, a ChatGPT-powered chat bot that provides question-answering support in educational spaces such as online forums and classrooms [41]. In related work, Castleman and Turkcan created three intelligent tutors using GPT-4, each with a different level of access to a knowledge base [6]. Other studies have applied LLMs to assessment and feedback tasks. Mizumoto and Eguchi used ChatGPT to score essays written by second language learners [28]; Lin et al. com-pared multiple GPT models in the task of providing feedback on tutor responses [22]; and Liu et al. assessed GPT 3.5 for evaluating the helpfulness of peer

*EDM-AIED 2025: Workshop on Epistemics and Decision-Making in AI-Supported Education. Co-located with AIED 2025, July 26, 2025, Palermo, Italy.*
*Corresponding Author
✉ langdon.holmes@vanderbilt.edu (L. Holmes); acoscia6@gatech.edu (A. Coscia); scott.crossley@vanderbilt.edu (S. A. Crossley); choijoonsuh@gmail.com (J. S. Choi); wesley.g.morris@vanderbilt.edu (W. Morris)
🆔 0000-0003-4338-4609 (L. Holmes); 0000-0002-0429-9295 (A. Coscia); 0000-0002-5148-0273 (S. A. Crossley); 0000-0002-7732-0266 (J. S. Choi); 0000-0001-6316-6479 (W. Morris)

review comments [23]. Finally, Singhal et al. used multiple GPT models to deidentify student writing and protect student privacy [38].

The above applications of general purpose LLMs to educational contexts make use of a prompt to adapt the LLM's behavior. However, researchers rarely provide adequate information about how prompts were developed and evaluated. Reports vary from a brief mention of the objectives that the prompt was designed to accommodate [28] to a more detailed description of multiple prompting strategies and how these strategies were meant to address the specific needs of the learning context [6]. Many researchers also provide the full prompt text in their report [22, 38, 41]. How-ever, few if any researchers describe the formal process by which prompt performance was evaluated. The general approach to developing LLM-powered interventions often relies on ad-hoc methods and iterative refinement rather than systematic evaluation. While educational technology studies frequently cite prompt engineering as part of their methodology, they rarely report on specific prompt design choices or their effects on task performance. This makes it difficult to build upon previous work or to determine which prompting strategies are most effective for different educational contexts.

This study presents a systematic approach to prompt design and evaluation in the context of a structured dialogue activity that takes place within an intelligent text webapp. Our approach addresses a critical decision-making challenge in AI-supported education: how to optimize prompts for the generation of learning content that is both personalized and pedagogically aligned. The study demonstrates (1) how prompt patterns can be leveraged to create more effective educational interactions with LLMs and (2) how the performance of different prompts can be evaluated in a replicable and resource efficient manner.

## 1.1. Prompt Design Patterns

A prompt is defined as an input to a generative LLM that is used to guide its output [12, 36]. A prompt generally comprises a set of specific instructions pertaining to the target task, and some contextual information that can help the LLM better under-stand the task. A prompt template is a customizable, structured blueprint that can be used to generate prompts [37]. Prompt templates contain slots that can be filled in at the time of prompt generation. For example, in the template "Generate a {{question_type}}", the question type is a variable that will be replaced by a target string at the time of prompt generation. Prompt templates explicitly define which elements are static across prompts and which should be replaced by situation-specific inputs.

Prompt design patterns provide reusable solutions to common problems encountered when engineering prompts for large language models (LLMs). Like software design patterns [10], prompt patterns codify proven approaches that can be adapted to different contexts while solving recurring challenges. White et al. present a comprehensive catalog of prompt patterns, organizing them into categories including Input Semantics, Output Customization, Error Identification, Prompt Improvement, and Interaction patterns [43]. Within the White et al. catalog, several prompt patterns are particularly relevant for educational applications:

For example, the Persona pattern can establish a specific role or character for the LLM, such as a tutor or coach, providing a consistent framework for generating responses. For example, the prompt "You are an expert mathematics tutor who specializes in explaining complex concepts using simple analogies." might be effective in educational contexts for maintaining an appropriate tone and expertise level for the LLM [23]. The Cognitive Verifier pattern structures the LLM's analysis process by breaking down complex questions into component parts before generating a response. This pattern aligns with established pedagogical techniques for scaffolding learning [49]. For example, "Before explaining the historical event, analyze (1) the key actors involved (2) the underlying societal conditions, and (3) the immediate trigger events, then use these elements to construct your explanation." The Context Manager pattern helps maintain focused educational dialogue by explicitly defining the scope of the conversation and relevant contextual elements. For example, "When discussing this physics problem, consider only classical mechanics principles. Do not introduce quantum mechanics concepts unless explicitly requested." The Alternative Approaches pattern encourages the LLM to consider multiple ways to accomplish a given task or solve a problem and present these to the user. For example, "Explain three different methods for solving quadratic equations." This pattern could help students to engage in comparative cognitive processes that might engage their critical faculties.

These prompt patterns can be combined to create more sophisticated prompt templates. For example, the Persona pattern might be combined with Context Manager to create a tutor that both maintains appropriate pedagogical stance and keeps the discussion focused on relevant topics. However, the effectiveness of such combinations in educational contexts requires systematic evaluation.

### 1.2. Evaluating LLM Outputs

With the proliferation of LLMs and prompt engineering for education, principled evaluations of LLM outputs have become critical. Two primary approaches have emerged for assessing LLM performance: metric-based evaluation and comparative judgment methods. Metric-based evaluation offers a granular approach to analysis, as demonstrated by Srivastava et al., who operationalized hundreds of dimensions for assessing LLM tasks and outcomes [40]. These dimensions span technical considerations like accuracy and latency, as well as human factors including agreeableness and cohesion. While this approach provides a detailed lens for examining LLM outputs, it can struggle with the complexity of holistic assessment. An alternative evaluation strategy is a comparative judgment method. The Chatbot Arena [48] exemplifies this approach by presenting users with outputs from different LLMs and asking them to select a preferred response. This method constructs a leaderboard based on user preferences, optimizing for a holistic assessment rather than predefined criteria [19]. Such an approach proves particularly valuable when evaluating complex constructs like LLM output quality, which often resist decomposition into discrete rubric elements.

Despite extensive work on LLM evaluation, principled prompt evaluation is under-studied. Conceptually, assessing LLM performance and evaluating prompts are equivalent, differing only in whether the prompt or the LLM is the primary object of manipulation. Researchers have recently called for more comprehensive approaches that simultaneously evaluate both LLMs and prompts, recognizing the sensitivity of LLM performance to prompting strategies [27]. In practice, prompt evaluation occurs most frequently through implicit, iterative prompt refinement [24]. The process often naturally occurs as humans encounter failure cases, such as needing to provide examples as "hints" to the model to handle edge-cases, a form of in-context learning for LLMs [45]. For example, Denny et al. show how natural language prompt engineering is easily adopted by novice programmers to write better prompts for models that generate code, such as Copilot [8]. However, this type of iterative design does not provide the same quality of evidence as a rigorous evaluation process. One problem is that humans can struggle to evaluate prompts at scale, often expecting LLMs to behave like humans and overgeneralizing results [47].

The current landscape of prompt evaluation research reveals a critical need for more structured, systematic methods of assessment. Despite the sophisticated approaches to LLM output evaluation, prompt evaluation remains largely unexplored. As educational applications using LLMs proliferate, developing robust, scalable methods for prompt evaluation will be crucial to designing effective and personalized learning interventions.

### 1.3. Current work

This study presents a systematic, generalizable approach to prompt evaluation that can be adopted across diverse learning contexts. We demonstrate our methodology within a specific reading comprehension dialogue system, but our primary contribution is the tournament-style evaluation framework itself. To illustrate this approach, we designed five prompt templates using established prompt patterns and adult learning theory. We then compared them against a Baseline template through a tournament-style evaluation with eight judges, assessing the quality of follow-up questions generated by an LLM using each template.

**RQ.** What does a tournament-style prompt evaluation process reveal about the performance of different prompting strategies in an educational dialogue?

### 1.4. Context

To demonstrate our approach in a real pedagogical context, we evaluate prompts designed for a structured dialogue activity. The Strategic Thinking and Interactive Reading Support (STAIRS) dialogue occurs within an intelligent text framework supporting students' reading comprehension and foundational reading skills. Intelligent texts created with this framework are interactive web apps that include text content augmented with a suite of learning activities. One such activity is the end-of-page summary, which requires students to write a summary of what they have read at the end of each page. These summaries are automatically evaluated across multiple dimensions using a scoring pipeline that combines multiple natural language processing (NLP) approaches.

The summary scoring system evaluates three key criteria: content, relevance, and language borrowing. Content scores assess whether the summary includes key ideas and details from the text. A fine-tuned Longformer model [1] evaluates the content dimension, with preliminary assessments [30] indicating that the model explains 79% of the content variance. Relevance is measured through semantic similarity between the source text and summary using a text embedding model [all-MiniLM-L6-v2, 33]. Language borrowing (also known as containment [4]) is analyzed using a Spacy-based pipeline [15] that examines trigram overlap between the source text and summary [4].

When a summary does not pass the content score threshold, STAIRS is activated to support the student's reading. The system first identifies an appropriate section of the text for the student to review. This decision is based on how long each section was visible on the student's screen relative to the word count of the section, whether the student had already engaged in an activity related to the section, and the semantic similarity between each section and the student's summary. The algorithm is intended to identify a section of the text with which the student engaged less deeply.

Using Llama 3 [9], STAIRS then generates a question for the selected passage based on self-explanation reading training [SERT, 26]. This framework is designed to support students' epistemic progression—the gradual development and refinement of their knowledge structures and understanding [11]. Epistemic progression involves not just accumulating information but advancing through qualitatively different ways of knowing and reasoning about content. Through SERT questioning, learners are prompted to articulate connections between textual information and their existing knowledge frameworks, identify relationships between concepts, and engage in metacognitive reflection about their comprehension processes. The question is randomly selected from five possible SERT question types: logic, bridging, prediction, elaboration, or paraphrasing. After the student re-reads the section and responds to the SERT question, STAIRS generates a follow-up question to deepen engagement with the text. After the learner responds to this follow-up question, they are directed to revise their summary.

## 2. Methods

We developed LLM prompts by populating a prompt template with a set of input data. The input data is sourced from authentic user interactions with STAIRS in the intelligent text framework. This data was sourced from three separate intelligent text deployments, each of which was based on a different textbook: 93 interactions are from Prolific crowd workers reading an economics text, 17 are from university students reading a psychology text, and 9 are from college students reading a programming text.

## 2.1. Prompt Templates

Our study evaluates six prompt templates: one Baseline template that has been used in previous intelligent text deployments and five templates that were designed using the prompt patterns described by White et al. [43]. The templates were crafted to generate follow-up questions after a learner's initial response to a Self-Explanation Reading Training (SERT) question.

Our approach to prompt design was grounded in principles of learning theory, with special emphasis on adult learners. Constructivist learning theory emphasizes that learners actively construct new knowledge by connecting it with their prior experiences [32], which aligns with research showing that adult learners have accumulated diverse experiences that make them more heterogeneous than adolescent learners [14, 39]. Similarly, Knowles' theory of andragogy [20] emphasizes that adult learners benefit from greater autonomy and agency in their learning process [34, 39]. Prompts were particularly influenced by theories of self-directed learning (SDL), which emphasize learner autonomy and helping students to develop a better understanding of their learning preferences [7, 17, 39]. The implementation of SDL in the prompts was guided by research on scaffolded learning conversations [44] and Vygotsky's zone of proximal development [42], which emphasizes the role of guided support in helping learners achieve tasks just beyond their current capabilities. This perspective aligns with research on metacognitive strategy development in adult education [50] and suggests that effective educational dialogue should support learner autonomy while providing structured opportunities for reflection and strategy development [26]. As a result, we designed the prompt templates in this study to emphasize learning and reading strategy development as well as scaffolding and meta-reflection.

Excluding the Baseline prompt template (374 words), all prompt templates were of a similar length (min =117 words, max = 192 words). Templates were designed to make use of prefix caching, which is an optimization strategy that can speed up response time by saving intermediate states of the LLM. The SERT question and learner response were different for every request to the LLM, so these elements were included at the end of the prompt using Llama's chat format. Static text and variable inputs that were repeated across multiple requests were included at the beginning of the prompt, which allowed our model serving software [vLLM, 21], to cache these prefixes. All templates used the same set of variable inputs: the title of the textbook, a one-paragraph description of the textbook, the full-text of the passage being discussed (between one and three paragraphs), the initial SERT question, and the learner's response to the initial SERT question. In total, six prompt templates were tested. The full prompt templates, as well as the prompt tournament annotation instructions, are available at https://osf.io/4jus9/?view_only=1413f65f0adb465a8ac5afbbd3190ebb.

**The Baseline** prompt was adapted from other prompts used within the intelligent text framework. While it was not designed with specific prompt patterns in mind, analysis revealed several key pattern implementations. The Persona pattern was employed through the establishment of a "reading support agent" role with clearly defined behavioral traits ("factual and concise") and explicit limitations ("You do not complete assignments"). The Context Manager pattern was implemented through multiple layers of control, including high-level principles based on Bloom's Taxonomy ("encourage users to engage with the text at different cognitive levels"), specific interaction guidelines ("concise responses", "factual and clear"), and explicit content boundaries through the text description and excerpt. The Cognitive Verifier pattern appeared in the prompt's approach to topic verification, instructing the LLM to "redirect the user to the topic by asking a question" if their message strayed from the specified text. The prompt included repeated emphasis on key constraints (e.g., "do not provide any form of summaries or overviews"), which was an attempt to maintain consistent LLM behavior through repetition rather than more sophisticated pattern implementation.

**The Socratic Guide** emphasized self-directed learning and critical thinking through Socratic questioning, aligning with Knowles' principles of adult learning [20]. This template employs the Persona pattern by establishing an "expert adult education facilitator specializing in Socratic questioning" role with clearly defined responsibilities. The Context Manager pattern is implemented through explicit scoping of the interaction around Socratic principles, including the directive to ask questions that "encourage metacognitive reflection" and to "help [students] make connections between their prior knowledge and the text." The template maintains tight control over the dialogue structure by explicitly defining the sequence: "You will ask the learner one self-explanation reading training question, the learner will respond, then you will ask a follow-up question."

**The Scaffolding Expert** incorporated Vygotsky's Zone of Proximal Development [42], primarily utilizing the Cognitive Verifier pattern through a structured three-step analysis process. This process analyzed three aspects: 1). The key concepts in the passage that the learner needs to understand, 2). The level of understanding demonstrated in the learner's response, and 3). The potential gaps or misconceptions in the learner's understanding. The Context Manager pattern was implemented through explicit scoping statements that direct the LLM to "target the most critical gap or opportunity for deeper understanding" and "use scaffolding principles to build from their current understanding."

**The Connection Builder** focused on constructivist learning principles, employed in the Context Manager pattern by explicitly defining the scope of connections the LLM should pursue. It did this by instructing the LLM to "help the learner connect ideas within the text" and "encourage them to draw on their personal experience or prior knowledge." The Alternative Approaches pattern was employed through directions to explore different types of connections while maintaining focus on "deepening their understanding through connecting different concepts or experiences." The template provided specific guidelines for handling off-topic responses: "If the learner response does not address the initial SERT question, then you should ask a question that will help steer the learning session back on-topic."

**The Strategic Reader Coach** emphasized metacognitive strategies and self-directed learning. It utilized the Persona pattern by establishing a "reading strategy coach" role with specific coaching objectives. The Context Manager pattern was implemented through explicit focus on strategic reading skills, directing the LLM to generate questions that "prompt the user to reflect on their reading strategy" and "help them identify key relationships in the text." The template maintained pedagogical focus through specific directives to "avoid suggesting specific interpretations" while encouraging metacognitive engagement.

**The Comprehension Monitor** incorporated self-regulated learning principles. It utilized the Cognitive Verifier pattern through structured analysis of "the main ideas of the passage," "the specific concepts addressed in the user's response," and "areas where deeper processing might be beneficial." The Context Manager pattern was implemented through explicit instructions to "help the user to evaluate their understanding" and "model a strategy that the user can use to monitor their comprehension." The template provided specific guidance for non-responsive answers: "If the user response does not address the initial SERT question, you may ask the user to try again."

### 2.2. Prompt Tournament

The prompt tournament presented judges with pairs of follow-up questions to initial STAIRS responses and asked judges to choose their preferred output. This is an example of a comparative judgment task, which asks judges to make a preference decision between multiple options rather than assigning an absolute score to a single sample at a time [19]. Each follow-up question was generated using one of the six prompt templates described above. The tournament used the Glicko2 rating system [13] to rank prompts in terms of their probability of producing follow-up questions that would be preferred over those generated by other prompts. In addition to the overall prompt ranking, the Glicko algorithm also produces a Bradley-Terry win probability for all pairwise comparisons between prompts, which is the estimated probability that one prompt's output will be preferred over another.

The tournament was conducted using Prodigy, a commercial annotation software developed by Explosion AI [29]. The annotator interface included annotation instructions, the conversation dialogue from real user data, and two follow-up questions generated using different prompts. Additional metadata of the user interaction was also provided. Annotators were asked to select which follow-up question they prefer using radio buttons. A "skip" button was available to annotators if they felt unable to make a preference assessment. This happened, for example, if the two follow-up questions were equivalent, or if both follow-up questions were unacceptable.

Eight judges participated in the tournament, contributing between 13-54 decisions (mean=30.42, SD=14.90) each. Judges were asked to complete 30 decisions over the course of two weeks. One judge was a faculty member, four judges were doctoral students, and three judges were undergraduate students. All were members of the intelligent text development team and were therefore familiar with both the specific learning context and educational technology more broadly. In total, judges made 213 preference decisions.

**Rubric.** To evaluate the quality of generated follow-up questions, we developed a three-part rubric focusing on format, dialogue support, and appropriateness for adult learners. The rubric was developed through an iterative process involving the intelligent text development team, with the goal of identifying questions that would effectively support learner engagement and comprehension.

The format criteria emphasized that responses should be presented as direct questions, allowing only brief supportive statements when contextually appropriate. This requirement ensures consistency in the interaction style and maintains focus on the dialogic nature of the reading support. Responses that include unnecessary preface text or procedural explanations (e.g., "Here is a follow-up question:") were considered less effective. The dialogue support criteria assessed how well the follow-up question built on both the initial SERT question and the learner's response. High-quality responses encourage discussion and deeper thinking about the text. Importantly, the rubric includes specific guidance for handling low-effort learner responses, preferring approaches that acknowledge lack of engagement and attempt to rebuild interest rather than simply repeating the initial question. The appropriateness criteria evaluate whether responses treat learners with appropriate respect and sophistication. These criteria favor questions that encourage connections between the text and learners' prior experiences, as well as questions that promote metacognitive reflection on reading strategies. This aligns with learning theories that emphasize self-directed learning and the importance of connecting new knowledge with existing experience during learning.

Judges were trained on the rubric using example pairs of responses that highlighted key distinctions in question quality. When evaluating response pairs, judges were instructed to consider all three rubric components while selecting the more effective response. This holistic approach, rather than a numerical scoring system, was chosen to better capture the nuanced differences in response quality that emerge from the interaction of multiple criteria.

**Rating System.** As mentioned earlier, the tournament employed the Glicko2 rating system [13], which provides an approach to ranking competitors based on paired comparisons. The system maintains three parameters for each competitor: a rating value ($\mu$), a rating deviation (RD) that indicates the reliability of the rating, and a volatility score ($\sigma$) that measures the degree of expected fluctuation in the rating.

Our prompt tournament was designed to maximize information gain by prioritizing comparisons between the most successful prompts at each time step. After each round of comparisons, the system identifies the two prompts with the highest ratings and schedules them for additional head-to-head evaluation. This adaptive sampling strategy helps to efficiently identify the strongest prompt by focusing evaluation effort on discriminating between the most promising candidates. Additionally, the system accounts for rating volatility through the RD parameter, allowing it to make more informed rating adjustments when judges show consistent agreement versus when their evaluations display more variability. This approach provides a framework for identifying the most effective prompt while making efficient use of annotator labor and accounting for uncertainties in both prompt performance and judge reliability.

## 3. Results

The prompt tournament results revealed clear differences in the effectiveness of different prompt templates for generating follow-up questions. We found that certain combinations of prompt patterns, particularly those emphasizing persona and context management, outperform others for generating follow-up questions. Our results further demonstrated that prompts grounded in learning theory produced more effective follow-up questions when combined with appropriate prompt engineering patterns.

The Strategic Reading Coach template emerged as the strongest performer, with consistently high win probability estimates against all other templates. It demonstrated an estimated win probability of 81% against the second-best performing prompt, the Scaffolding Expert, and win probabilities of at least 90% compared to all other prompts. The estimated win probabilities and number of trials are reported in Table 1. The number of trials varies between matchups, from 0 to 91, because we designed the tournament as an adaptive comparative judgment task that prioritizes collecting information about the two most effective prompts at each time step. This was done to make the most effective use of annotator labor.

**Table 1**
Results of Prompt Tournament

| Prompt A | Prompt B | Prob A > B | Trials |
|---|---|---|---|
| SRC | SE | 0.81 | 55 |
| SRC | Baseline | 1.00 | 14 |
| SRC | CM | 1.00 | 1 |
| SRC | SG | 0.98 | 2 |
| SRC | CB | 0.94 | 3 |
| SE | Baseline | 1.00 | 6 |
| SE | CM | 1.00 | 0 |
| SE | SG | 0.96 | 9 |
| SE | CB | 0.91 | 11 |
| Baseline | CM | 1.00 | 91 |
| Baseline | SG | 0.85 | 23 |
| Baseline | CB | 0.77 | 2 |
| CM | SG | 0.55 | 0 |
| CM | CB | 0.56 | 0 |

*Note:* SRC = Strategic Reading Coach, SE = Scaffolding Expert, CM = Comprehension Monitor, SG = Socratic Guide, CB = Connection Builder

The Scaffolding Expert template emerged as the second-best performer overall, also showing strong performance across its matchups. Against the Baseline template, it achieved a 100% win probability. The Scaffolding Expert template also performed strongly against the Socratic Guide and Connection Builder, with win probabilities of 96% and 91% respectively.

The large margins of victory between templates suggest that the design differences had substantial impact on output quality. The Baseline template, despite not being explicitly designed with prompt patterns, performed relatively well, placing third overall. During the early phases of the tournament, the Baseline template was outperforming other prompts, which explains the high number of trials for the comparison between the Baseline and Comprehension Monitor templates. This observation indicates that some of the Baseline template's existing features, such as the emphasis on Bloom's Taxonomy and metacognition, align well with the pedagogical aims of the follow-up question.

The results suggest that certain prompt patterns were particularly effective when combined. The Strategic Reading Coach's combination of the Persona Pattern and Context Manager Pattern produced consistently strong results, suggesting these patterns work well together for generating follow-up questions. Similarly, the Scaffolding Expert's use of the Cognitive Verifier Pattern and Reflection Pattern proved effective, indicating the value of structured analysis before question generation. In contrast, templates using the Context Manager Pattern (Comprehension Monitor) or Alternative Approaches Pattern (Connection Builder) performed less well, suggesting these patterns were less suited for generating follow-up questions in this specific context.

## 4. Discussion

The goal of the research reported in this study is to showcase methods to optimize prompts for specific pedagogical objectives. As LLMs become increasingly integrated into educational technologies, structured approaches to prompt engineering and evaluation will be essential for creating effective, theoretically grounded learning experiences. The results of our prompt tournament provided clear and quantitative evidence about the most effective prompt template in a specific educational application. Importantly, the prompt tournament methodology used in this study is highly adaptable to other educational contexts. Additionally, the approach helped us glean insights into the effectiveness of different prompt engineering approaches for educational dialogue.

The Strategic Reading Coach (SRC) template emerged as the clear winner, demonstrating superior performance across all pairwise comparisons with win probabilities ranging from 81% to 100%. This strong performance can be attributed to several key factors revealed through our analysis. The SRC template's combination of the Persona and Context Manager patterns proved particularly effective for generating follow-up questions that maintain pedagogical focus while adapting to learner responses. The success of this combination suggests that establishing a consistent tutoring persona while simultaneously refining questions based on learner input creates a more effective dialogue structure than approaches that rely on single patterns.

An unexpected finding was the relatively strong performance of the Baseline template, which placed third overall despite not being explicitly designed with modern prompt patterns. This suggests that some fundamental principles of educational dialogue, such as promoting metacognition and following Bloom's Taxonomy, remain effective regardless of how they are implemented in prompt engineering. However, the substantial performance gap between the Baseline and the top two templates indicates that thoughtful application of prompt patterns can significantly improve output quality.

The tournament results also highlighted areas where several prompt engineering approaches fell short. The poor performance of the Connection Builder template, despite its grounding in constructivist learning theory, suggests that theoretical alignment alone does not guarantee effective prompt performance. This finding highlights the importance of empirical evaluation in prompt engineering.

Our findings demonstrate the value of systematic prompt evaluation in educational applications, moving beyond the common practice of ad-hoc prompt engineering. The tournament methodology proved effective for identifying meaningful differences between prompt templates. The current study, however, is not without limitations. First, our evaluation focused solely on follow-up questions within a specific reading comprehension context. While the tournament methodology can be adapted for other educational applications, the specific findings about prompt effectiveness might not generalize beyond this context. Second, our analysis used a single LLM (Llama 3) due to deployment constraints. Different models might respond differently to the same prompt patterns, suggesting a need for cross-model evaluation in future work. One advantage of the prompt tournament design is that different combinations of prompt and LLM can be evaluated simultaneously. This could prove important when deciding which LLM to use in a given application, as the same prompt might perform better or worse when used with a different LLM [27].

## 5. Future Work

This work represents an initial step toward more systematic design and evaluation of prompts for educational applications of large language models. While our tournament methodology focuses on comparing prompts for generating follow-up questions in reading comprehension, the approach could be adapted for evaluating prompts across a range of educational tasks and contexts. Future research should explore how different combinations of prompt patterns and LLMs perform across various pedagogical goals. Additionally, future work might explore methodological improvements to the prompt design process. For example, interactive interfaces like PromptMaker [18] and ConstitutionMaker [31] have been developed to create a structured prompt design process. The artifacts produced by this process could help to make prompt engineering practices more describable and reproducible in educational technology research.

## Acknowledgements

This material is based upon work supported by the National Science Foundation under Grant 2112532 Any opinions, findings, and conclusions or recommendations expressed in this material are those of the author(s) and do not necessarily reflect the views of the National Science Foundation.

## Declaration on Generative AI

The author(s) have not employed any Generative AI tools.